\documentclass[10pt,twocolumn,letterpaper]{article}

\usepackage{cvpr}
\usepackage{times}
\usepackage{epsfig}
\usepackage{graphicx}
\usepackage{amsmath}
\usepackage{amssymb}
\usepackage{tabularx}
\usepackage{setspace} 
\usepackage{booktabs}
\usepackage{chngpage}

\usepackage[font=small,labelfont=bf, justification=justified,
format=plain]{caption}
\usepackage[pagebackref=true,breaklinks=true,letterpaper=true,colorlinks,bookmarks=false]{hyperref}
\cvprfinalcopy

\def\cvprPaperID{4174} 

\ifcvprfinal\fi

\begin{document}
\title{Attention-based Few-Shot Person Re-identification Using Meta Learning}


\author{Alireza Rahimpour  \hspace{2cm}    Hairong Qi \\			
Department of Electrical Engineering and Computer Science\\
University of Tennessee, Knoxville, TN, USA\\
}

\maketitle
	\begin{abstract}
	In this paper, we investigate the challenging task of person re-identification from a new perspective and propose an end-to-end attention-based architecture for few-shot re-identification through meta-learning. 
	The motivation for this task lies in the fact that humans, can usually identify another person after just seeing that given person a few times (or even once) by attending to their memory. On the other hand, the unique nature of the person re-identification problem, i.e., only few examples exist per identity and new identities always appearing during testing, calls for a few-shot learning architecture with the capacity of handling new identities. Hence, we frame the problem within a meta-learning setting, where a neural network based `meta-learner' is trained to optimize a `learner' --- an attention-based matching function. 
	Another challenge of the person re-identification problem is the small inter-class difference between different identities and large intra-class difference of the same identity. In order to increase the discriminative power of the model, we propose a new attention-based feature encoding scheme that takes into account the critical intra-view and cross-view relationship of images. We refer to the proposed Attention-based Re-identification Meta-learning model as ARM. 
	Extensive evaluations demonstrate the advantages of ARM as compared to the state-of-the-art on the challenging PRID2011, CUHK01, CUHK03 and Market1501 datasets.
\end{abstract}

\section{Introduction}
\begin{figure}[t!]
	\centering
	\includegraphics[width=0.8\linewidth]{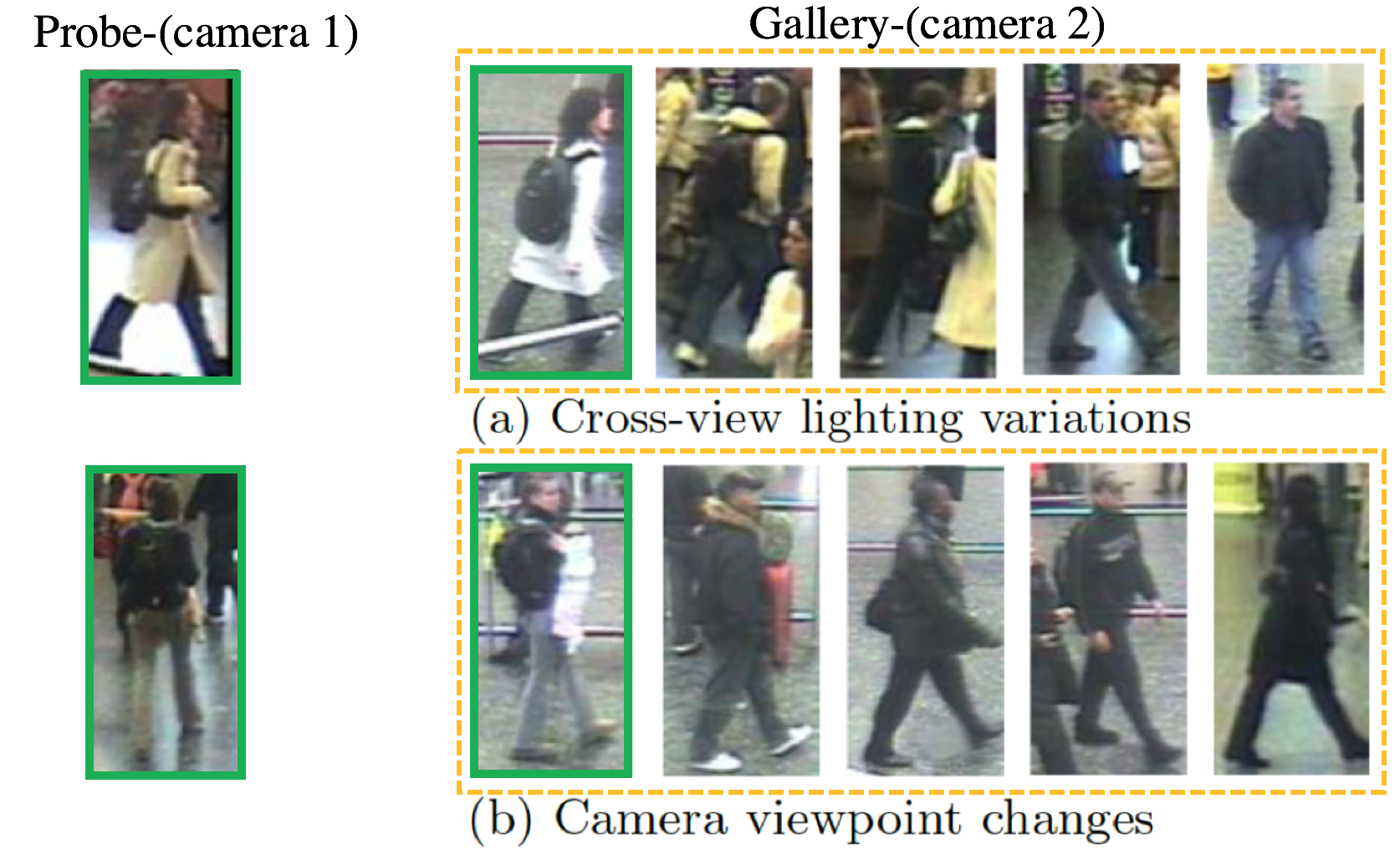}
	\caption{Two examples of camera view variation challenge. The images of the same person (the ones with green border in each row) look different in two different camera views. Also, images from different people in the gallery may look the same. Our model is able to handle these challenging situations.}
	\label{fig:fig1}
\end{figure}

Recently, person re-identification has gained increasing research interest in the computer vision community due to its importance in multi-camera surveillance systems. In person re-identification, the goal is matching people across non-overlapping camera views at different times. Despite all the research efforts, person re-identification remains a challenging problem since a person's appearance can vary significantly when large variations in view angle, illumination, human pose, background clutter and occlusion are involved, as shown in Fig.~\ref{fig:fig1}. 
To address these difficulties, several approaches have been proposed in recent years. These algorithms either focus on learning more discriminative metrics for comparing person images or extracting more representative visual features. Specifically, inspired by the success of deep neural networks in many computer vision tasks, deep architectures have been widely used for person re-identification and achieved state-of-the-art results (e.g., \cite{cheng2016person,varior2016learning, ahmed2015improved, zhang2016learning}). 
However, there are still challenging issues remaining in solving the person re-identification problem.

\textbf{The first challenge} is the lack of examples per identity as well as difficulties in training these deep learning based models. 
Most recent supervised deep learning based approaches consider re-identification as a classification problem and have demonstrated better performance \cite{zheng2016person} than the matching-based models (e.g., Siamese \cite{yi2014deep}).  However, these methods need lots of labeled training data per ID which might not be feasible for real-world deployment with data exhibiting the characteristic of having many different classes but few examples per class. In fact, in most of the existing re-identification datasets the number of images per identity for each view is on average less than $5$ (e.g., each identity has $2$, $4.8$ and $3.6$ images on average for each view in CUHK01, CUHK02 and Market1501, respectively).

Another trend is to consider re-identification as a matching problem which would address the lack of example issue. Recent development in this direction includes the triplet loss and its variants \cite{cheng2016person,su2016deep,ding2015deep,wangjoint}.
However, training of these triplet-based networks is challenging and requires specific algorithms (e.g., hard negative mining \cite{schroff2015facenet}) for selecting the triplets and the margin. Improper selection can in practice lead to bad local minima early on in training or very slow convergence. In addition, since the triplet loss based networks consider the pairwise relationship between images as triplets only during training, they suffer from an imperfect generalization capability in testing \cite{chen2017beyond}. Finally, semi-supervised and unsupervised methods have been proposed \cite{liu2017stepwise, zhao2013unsupervised} to address the need for labeled data, but unfortunately their performance has not been quite in par with supervised methods. 

\textbf{The second challenge} is how to model the relationship between all the gallery and probe images in the feature extraction process (in both training and testing). 
In fact, different individuals can share a similar appearance, and the appearance of the same person can be drastically different in different camera views. Fig.~\ref{fig:fig1} shows two instances of view variations challenge. For example, even though the gallery images of the second example (b) look the same, they actually belong to different identities. 
Therefore, cross-view and intra-view relationships play a critical role in person re-identification. 
Most existing approaches fail to define a structure that compares the probe image to the whole gallery images as a set when extracting the feature representation of each image. The performance of learned features using existing criterion based on pairwise similarity is still limited, because only point-to-point (i.e., image-to-image) relationships are mostly considered in a Siamese or triplet loss structure. 
Although classification-based approaches use a softmax layer where the relationship between images is implicitly considered, the feature of each image is still extracted independently of all the other images in the gallery.



To \textbf{address the first challenge}, we propose an attention-based model for few-shot person re-identification. 
The motivation lies in the fact that humans, even children, can usually generalize after seeing the same person only a couple of times. Moreover, few-shot re-identification would help alleviate data collection and annotation in large camera networks as we would not require large amount of labeled examples to attain reasonable performance. 
The proposed model is able to learn quickly from a few examples during training and is able to identify new people in the gallery in the testing stage. In order to deal with new classes in the gallery and to acquire quick knowledge from few examples during training, we exploit the idea of meta-learning. Meta-learning suggests framing the learning problem at two levels. The first level is the `learner' that learns a metric for performing the matching between each probe image and the gallery images. The second level is the `meta-learner' that guides the `learner' across all the matching processes.  
Unlike previous works, by using this framework, there is no need for any additional procedure for triplet or margin selection during training. 
Furthermore, since meta-learning optimizes an objective during training which directly reflects the person re-identification task during testing, the model has more generalization power to cope with unobserved identities without any changes to the network.

To \textbf{address the second challenge}, we propose an attention-based feature encoding structure, where we take into account the relationship among all the gallery images as well as that between all the gallery and probe images in the feature encoding process. The relationship between all the images inside the gallery is modeled using the proposed gallery encoder architecture. Similarly, the relationship between the probe image and the gallery set is modeled using the probe encoder architecture. 
The proposed feature encoding framework enhances the discriminative capability of feature representation by leveraging the cross-view and intra-view relationship between images and selectively propagating relevant contextual information throughout the network. In this way, compared to existing works, the proposed network can deal with the view-specific matching task more effectively. Moreover, the memory structure in the attention-based encoding framework helps the meta-learning algorithm to remember how to update the learner, thus facilitates the handling of new unseen identities in the gallery.  
We refer to the proposed Attention-based Re-identification Meta-learning model as ARM. 

The main contributions of this work are summarized as follows:
\begin{itemize}{
		\vspace{-0.2cm}		
		\item Introducing a novel end-to-end attention-based person re-identification framework which is able to perform few-shot learning by exploiting the concept of meta-learning. 
		
		\item Designing an attention-based gallery encoder to incorporate the intra-view and cross-view relationships between gallery images in order to generate more representative and discriminative features.}	
	
	\item Designing an attention-based probe encoder in order to model the cross-view relationship between the probe and gallery images. The proposed task-driven encoder attends to the encoded features of all the gallery images and incorporates their inter-relationship into feature representation of each probe image. 
	
\end{itemize}
This is an ongoing work which is being revised and more details will be provided soon.


{\small
	\vspace{0 cm}
	\bibliographystyle{ieee}
	\bibliography{egbib}
	\vspace*{\fill}
}

\vspace*{\fill}
\pagebreak
\appendix
\vspace*{\fill}

\end{document}